\documentclass[10pt,twocolumn,letterpaper]{article}

\usepackage{cvpr}
\usepackage{times}
\usepackage{epsfig}
\usepackage{graphicx}
\usepackage{amsmath}
\usepackage{amssymb}
\usepackage{multirow}
\usepackage{pifont}
\usepackage{booktabs}
\usepackage{epsfig}
\usepackage{booktabs,multirow}
\usepackage{graphics}
\usepackage{threeparttable}
\usepackage{color}
\usepackage[normalem]{ulem}
\usepackage{multirow}
\usepackage{float}
\usepackage{amsfonts}
\usepackage{bm}
\usepackage{subfig}
\usepackage{enumitem}
\usepackage[numbers]{natbib}
\usepackage{array}
\usepackage[table]{xcolor}
\usepackage{colortbl}

\usepackage{bm}
\usepackage{amsmath}

\newcolumntype{I}{!{\vrule width 1pt}}

\DeclareMathOperator*{\argmax}{argmax}

\makeatletter
\newcommand{\thickhline}{%
    \noalign {\ifnum 0=`}\fi \hrule height 1pt
    \futurelet \reserved@a \@xhline
}
\makeatother

\usepackage{caption}
\usepackage[pagebackref=true,breaklinks=true,letterpaper=true,colorlinks,bookmarks=false]{hyperref}

\usepackage[utf8]{inputenc}

\definecolor{bblue}{RGB}{0,30,95}
\definecolor{rred}{RGB}{190,0,0}
\definecolor{mygray}{gray}{.9}
\definecolor{ggray}{RGB}{127,127,127}

\usepackage{cleveref}
\crefname{section}{§}{§§}
\Crefname{section}{§}{§§}
\cvprfinalcopy 

\def\cvprPaperID{2791} 

\begin{document}

\title{Structured Scene Memory for Vision-Language Navigation}

\author{Hanqing Wang$^{1}$~, Wenguan Wang$^{2}$\thanks{Corresponding author: \textit{Wenguan Wang}.}~~, Wei Liang$^{1*}$, Caiming Xiong$^3$~, Jianbing Shen$^{1,4}$\\
\small{$^1$Beijing Institute of Technology~~~$^2$ETH Zurich~~~$^3$Salesforce Research~~~$^4$Inception Institute of Artificial Intelligence}\\
\small\url{https://github.com/HanqingWangAI/SSM-VLN}
}

\maketitle
\thispagestyle{empty}

\begin{abstract}
Recently, numerous algorithms have been developed to tackle the problem of vision-language navigation (VLN), \ie, entailing an agent to navigate 3D environments through following$_{\!}$ linguistic$_{\!}$ instructions.$_{\!}$ However,$_{\!}$ current$_{\!}$ VLN$_{\!}$ agents simply store their past experiences/observations as latent states in recurrent networks, failing to capture environment layouts and make long-term planning. To address these limitations,  we propose a crucial architecture, called Structured Scene Memory (SSM). It is compartmentalized enough to accurately memorize the percepts during navigation.  It also serves as a structured scene representation, which captures and disentangles visual and geometric cues in the environment.  SSM has a collect-read controller that adaptively collects information for supporting current decision making and mimics iterative algorithms for  long-range reasoning. As SSM provides a complete action space, \ie, all the navigable places on the map, a frontier-exploration based navigation decision making strategy is introduced to enable efficient and global planning. Experiment results on two VLN datasets (\ie, R2R and R4R) show that our method achieves state-of-the-art performance on several metrics.
\end{abstract}

\vspace{-3pt}
\section{Introduction}
\vspace{-2pt}
As a crucial step towards building intelligent embodied agents, autonomous navigation has long been studied in robotics. Since Anderson \etal\!~\cite{anderson2018vision} extended prior efforts\!~\cite{chen2011learning,mei2016listen} in instruction based navigation into photo-realistic simulated scenes\!~\cite{Matterport3D}, vision-language navigation (VLN) has recently attracted increasing attention in computer vision community. Towards the goal of enabling an agent to execute navigation instructions in 3D environments, current representative VLN methods made great advances in: \textbf{i)} developing more powerful learning paradigms\!~\cite{wang2018look,wang2019reinforced}; \textbf{ii)} exploring extra supervision signals from synthesized data\!~\cite{fried2018speaker,tan2019learning,fu2019counterfactual} or auxiliary tasks\!~\cite{wang2019reinforced,huang2019transferable,ma2019self,zhu2019vision}; \textbf{iii)} designing more efficient multi-modal embedding schemes\!~\cite{hu2019you,qi2020object,wang2020environment}; and \textbf{iv)} making more intelligent path planning~\cite{ke2019tactical,ma2019regretful,wang2020active}.

\begin{figure}[t]
	\begin{center}
		\includegraphics[width=\linewidth]{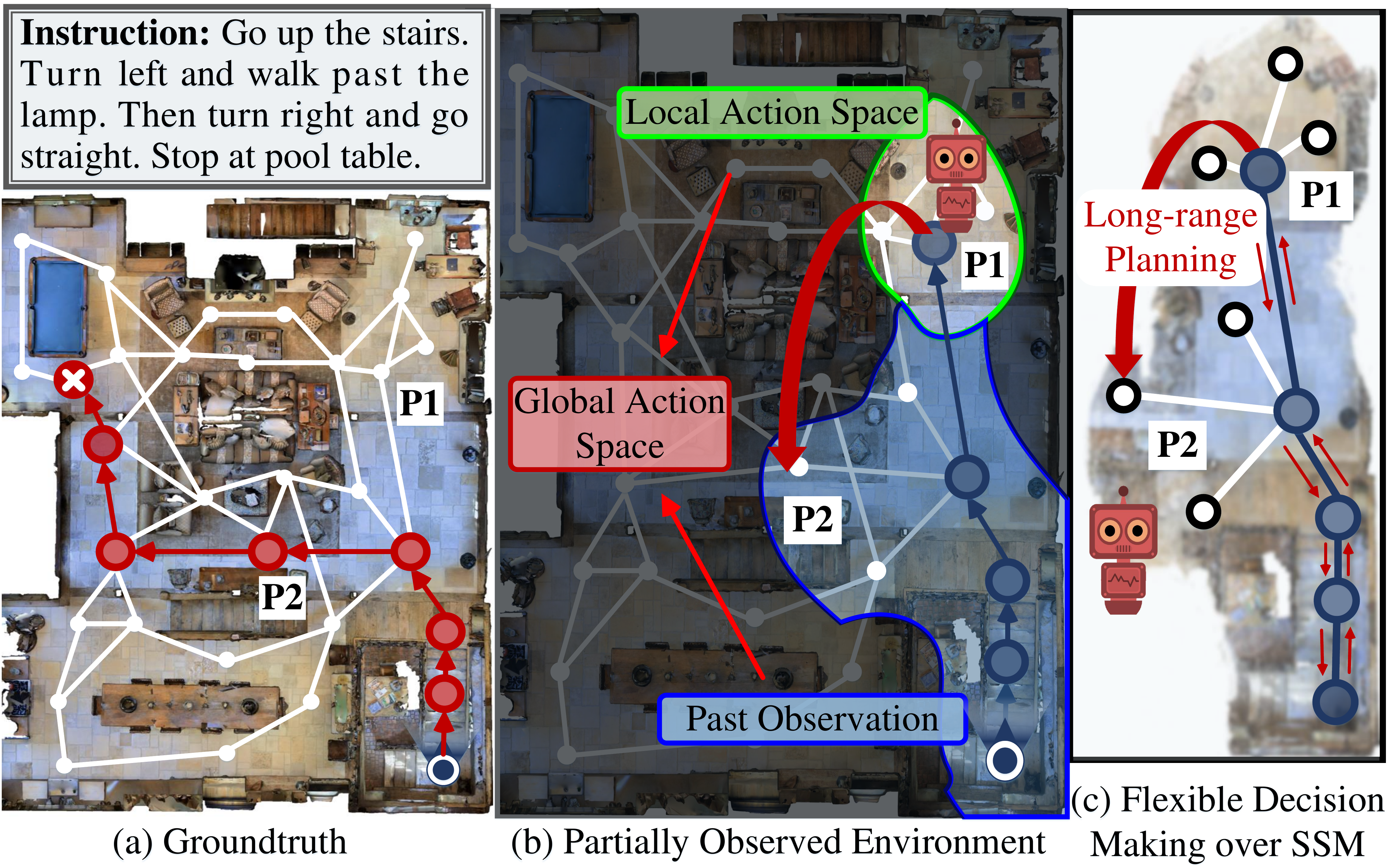}
	\end{center}
	\vspace{-16pt}
	\captionsetup{font=small}
	\caption{\small{In the partially observed environment (b), our agent builds SSM (c) for persistent memorization and topological scene representation. SSM supports long-range planning over the global action space, thus our agent can easily change its navigation direction (\ie, P1) by directly `jumping' to any previously visited position (\eg, P2), resulting in robust navigation.} }
	\label{fig:motivation}
	\vspace{-13pt}
\end{figure}

Despite these progresses, current VLN models typically follow\!~\cite{anderson2018vision} to tackle the task through a sequence-to-sequence (Seq2Seq) framework, which maps instructions and online observations to navigation actions. With such a design, the perceived information are embedded and mixed in the internal recurrent units. This prohibits the agent from directly accessing to its past observations and understanding the environment layout. Taking Fig.\!~\ref{fig:motivation} as an example. VLN agents execute instructions in a partially observed environment (Fig.\!~\ref{fig:motivation}(b)). As current VLN agents simply encode past perceptions (blue area in Fig.\!~\ref{fig:motivation}(b)) as hidden network states,  their decision-making is only limited within a local action space (\ie, currently navigable directions -- green area in Fig.\!~\ref{fig:motivation}(b)). Thus they tend to make sub-optimal navigation decisions, failing to perform path planning over the explored environment (\ie, global action space in Fig.\!~\ref{fig:motivation}(b)).


In robotics, map-building is long studied for facilitating path planning~\cite{thrun2002probabilistic,fuentes2015visual}. Inspired by classic simultaneous localization and mapping (SLAM) models~\cite{thrun2002probabilistic,hartley2003multiple}, recent efforts make use of deep learning techniques for constructing more efficient space representations\!~\cite{gupta2017cognitive,fang2019scene,chaplot2020learning,chaplot2020neural}.
On the other hand, making path planning in VLN, in essence, requires efficiently modeling long-term dependencies within online observation sequences. However, the states (\textit{internal} memory) in recurrent networks are latent and shown inherently unstable over long timescales~\cite{sukhbaatar2015end}. Alternatively, neural networks with \textit{external} memories provide a feasible solution, which uses a compartmentalized memory to preserve and recall long-term information and has been shown successful in language modeling\!~\cite{sukhbaatar2015end} and deep reinforcement learning (RL) agents in 3D environments~\cite{oh2016control,parisotto2017neural}.

Built off these two lines of previous efforts, we propose an essential, outside memory architecture, called Structured Scene Memory (SSM). SSM online collects and precisely stores the percepts during navigation (see Fig.\!~\ref{fig:motivation}(b-c)). It is graph-structured, yielding a topological representation of the environment. Thus SSM is able to perform long-term memorization and capture the environment layouts, allowing our agent to make efficient planning. In addition, SSM delivers a global action space -- all the navigable locations within the explored area, enabling flexible decision making. As shown in Fig.\!~\ref{fig:motivation}(c), when necessary,  our agent can simply move off its current navigation direction (\ie, P1) and travel to another previously visited far location (\ie, P2).


Specifically, SSM consists of nodes that embed visual information in explored locations and edges that represent geometric relations between connected locations. As both visual and geometric cues are captured and disentangled, instructions can be better grounded onto the visual world, \ie, correlating perception and action related descriptions with nodes and edges in SSM respectively. SSM is equipped with a collect-read controller, which adaptively reads content from the memory,  depending on current navigation context. The controller also mimics iterative algorithms for comprehensive information gathering and  long-range reasoning. Hence, SSM brings a global action space but its gradually increased scale will make policy learning harder. We propose a frontier-exploration based decision making strategy that addresses this issue elegantly and efficiently.

Extensive experiments on Room-to-Room  (R2R)\!~\cite{anderson2018vision} and Room-for-Room (R4R)\!~\cite{jain2019stay} datasets demonstrate the effectiveness of the full approach and core model designs.


\section{Related Work}

\noindent\textbf{Vision-Language Navigation (VLN).} Since the release of R2R dataset\!~\cite{anderson2018vision}, numerous VLN methods are proposed, which mainly focus on the use of deep multimodal learning to arrive at navigation policies. For instance, from imitation learning (IL)\!~\cite{anderson2018vision}, to the hybrid of model-free and model-based RL\!~\cite{wang2018look}, to the ensemble of IL and RL\!~\cite{wang2019reinforced}, different learning regimes are developed. For further boosting learning, some methods try to mine extra supervisory signals from synthesized samples\!~\cite{fried2018speaker,tan2019learning,fu2019counterfactual} or auxiliary tasks~\cite{wang2019reinforced,huang2019transferable,ma2019self,zhu2019vision}. Hence, for more intelligent path planning, the abilities of self-correction~\cite{ke2019tactical,ma2019regretful} and active exploration~\cite{wang2020active} are addressed. Some other recent studies also address environment-agnostic representation learning~\cite{wang2020environment}, fine-grained instruction grounding~\cite{hong2020sub,qi2020object,hong2020language}, web image-text paired data based self-pretraining~\cite{majumdar2020improving,hao2020prevalent}, or perform VLN in continuous environments~\cite{krantz2020navgraph}.


Existing VLN agents are built upon Seq2Seq models~\cite{anderson2018vision}, where past observations are encoded as the hidden states of recurrent units. Thus they struggle to precise memorization over long time lags and fail to explore environment layouts. In contrast, our agent is empowered with an external, structured memory for effective space representation and efficient memory operation. It provides a global action space and allows flexible decision making. A concurrent work\!~\cite{deng2020evolving} also addresses planning over topological space representation. In \cite{deng2020evolving}, visual and geometric information of the environment are mixed together in the scene map. But these two are disentangled in ours, facilitating instruction grounding. 
Hence, to avoid a huge action space, we propose a frontier-exploration based decision making strategy, with a navigation utility driven policy learning regime. However, \cite{deng2020evolving} requires a specifically designed planner with a subset of the decision space and trains the agent only  with IL. 



\begin{figure*}[t]
	\begin{center}
\vspace{-6pt}
		\includegraphics[width=\linewidth]{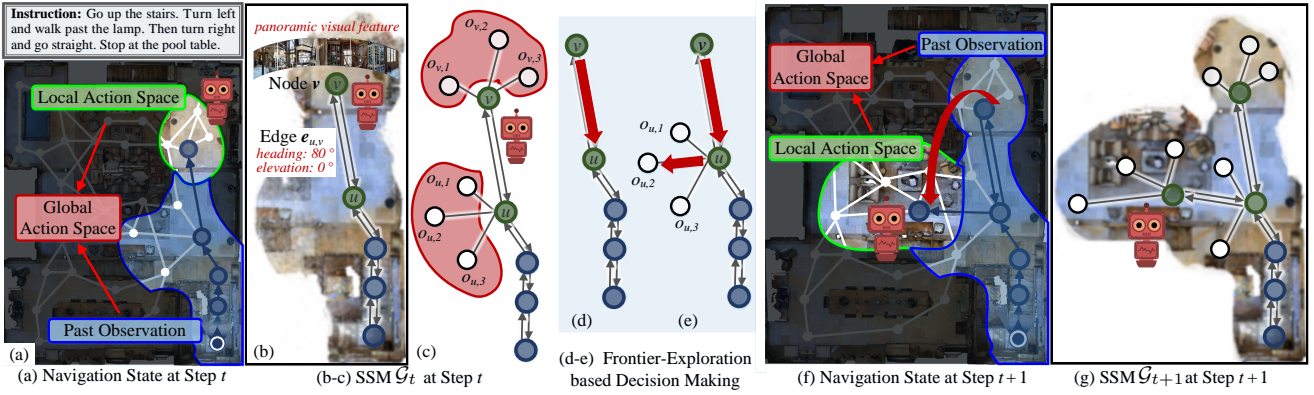}
    \put(-347,3){\footnotesize $\mathcal{G}_t$}
    \put(-66,3){\footnotesize $\mathcal{G}_{t+1}$}
	\end{center}
	\vspace{-17pt}
	\captionsetup{font=small}
	\caption{\small{Previous VLN agents are limited to a local action space, while SSM allows our agent to store past observations, explicitly model visual and geometric cues in environmental layout, and make robust VLN planning over the global action space. Please see \S\ref{sec:3} for details.} }
	\label{fig:overview}
	\vspace{-14pt}
\end{figure*}

\noindent\textbf{Neural Networks with External Memory.} Many tasks require capturing structures within sequential data, such as machine translation and question answering.
Though recurrent networks~\cite{hochreiter1997long} have been shown Turing-Complete~\cite{siegelmann1995computational} and applicable for sequential learning, their ability in  modeling long-term data structure is limited, as their memory are latent network states. Recent work~\cite{graves2014neural} further explores augmenting networks with outside memory, which allows for long-term memorization as well as explicit memory manipulation (such as read and write operations). Some representative ones show remarkable success in  language modeling~\cite{graves2014neural,weston2014memory,sukhbaatar2015end,bahdanau2014neural}, robot control~\cite{oh2016control,parisotto2017neural}, and meta-learning \cite{santoro2016meta}. Inspired by their novel ideas, we develop a structured, external memory architecture, which can store percepts during long-time exploration and capture environment layouts. With a collect-read controller that addresses adaptive context information gathering, we believe our memory design is well-suited to the performance of the VLN task.

\noindent\textbf{Space Representation and Path Planning in Navigation.} To perform autonomous navigation, it is crucial to humanize a robot into learning spatial environments and making planning. Classic strategies typically build space representation via SLAM \cite{thrun2002probabilistic,hartley2003multiple}, fusing  multi-modal signals, such as LIDAR, depth, or structure from motion. With the estimated maps, path planning is made \cite{kavraki1996probabilistic,lavalle2001rapidly,canny1988complexity} for traveling to goal locations. Recent efforts focus on learning navigation policies over spatial representations~\cite{gupta2017cognitive,parisotto2017neural,zhang2017neural}, topological maps~\cite{savinov2018semi,chen2019behavioral,wu2019bayesian,chaplot2020neural,chaplot2020learning}, or designing trainable  planners~\cite{lee2018gated,deng2020evolving}. As for VLN, current methods rarely make use of scene maps, and thus cannot make robust planning. Although some ones address better planning through rule-based beam search~\cite{fried2018speaker}, self-correction~\cite{ke2019tactical,ma2019regretful}, or active perception~\cite{wang2020active}, they still suffer from a local action space and latent memory. Instead, we explore and store scene representation through a structured memory, which provides a global action space and facilitates long-term reasoning. Hence, a frontier-exploration based decision making strategy is adopted to prevent the rapidly expanding action space from causing expensive learning and slow convergence.




\vspace{-2pt}
\section{Our Approach}\label{sec:3}
\vspace{-2pt}
\noindent\textbf{Task Setup and Notations.} Following the standard VLN setup~\cite{anderson2018vision}, an embodied agent is required to online navigate through environments to reach target locations, according to language instructions. To be able to focus on high-level planning, in~\cite{anderson2018vision}, the environment is assumed to be
viewed from a set of sparse points and the navigability between the points is given. Formally, let us denote by $\bm{X}\!=\!\{\bm{x}_l\}_{l\!}$ the textual embedding of the instruction with $|\bm{X}|$ words.
At each step $t$, the agent observes a panoramic view\!~\cite{fried2018speaker}, which is discretized into 36 single views (\ie, RGB images). Hence, there are $K_t$ navigable viewpoints $\{{o}_{t,k}\}_{k=1\!}^{K_t}$ that are reachable and visible. Each navigable viewpoint ${o}_{t,k}$ is associated with a visual feature $\bm{f}_{t,k}$ and orientation feature $\bm{r}_{t,k}=(\cos\theta_{t,k}, \sin\theta_{t,k}, \cos\phi_{t,k}, \sin\phi_{t,k})$, where $\theta$ and  $\phi$ are the angles of heading and elevation, respectively.

\noindent\textbf{Previous Solutions to VLN.} Previous VLN agents~\cite{anderson2018vision,tan2019learning}, in general, are built as a recurrent action selector based on received observations, instructions, and prior latent state. At each step $t$, the selector updates its internal state through:
\vspace{-2pt}
\begin{equation}\small
\begin{aligned}
\bm{h}_t = \text{LSTM}([\bm{X}, \bm{O}_{t}], \bm{h}_{t-1}),
\end{aligned}
\vspace{-3pt}
\label{eq:hidden-state}
\end{equation}
where $\bm{O}_{t}\!\!=\!\!\{\bm{o}_{t,k}\}_{k=1}^{K_t}\!\!=\!\!\{[\bm{f}_{t,k}, \bm{r}_{t,k}]\}_{k=1}^{K_t}$, and `$[\cdot]$' indicates the concatenation operation.

Current navigable viewpoints $\{{o}_{t,k}\}_{k=1}^{K_t}$ form the action space $\mathcal{A}_t$, from which the moving action $a_t$ is selected:
\vspace{-4pt}
\begin{equation}\small
\begin{aligned}
\!\!\!\!\!\!\!\!\!\!a_t = \argmax_{k} ~p(o_{t,k}).
\end{aligned}
\label{eq:decision-making}
\vspace{-6pt}
\end{equation}
Here $p(o_{t,k}) = \text{softmax}_k(\bm{o}_{t,k}^\top\bm{W}^p_o\bm{h}_t)$, where $\bm{W}^p_o$ is a learnable parameter matrix.  Defined in this way, $p(o_{t,k})$ returns the probability of the viewpoint ${o}_{t,k}$ to be acted on.


\noindent\textbf{Our Idea.}  With above brief descriptions of the core idea of existing VLN models, some limitations can be exemplified. First, they fail to explore environmental layouts. Second, their planning ability is limited to the latent memory $\bm{h}$, which cannot precisely store and directly recall their past experience, \ie, $\{\bm{O}_{1}, \cdots, \bm{O}_{t}\}$. Third, they suffer from a local action space $\mathcal{A}_t$, \ie, immediate observation $\{{o}_{t,k}\}_{k=1}^{K_t}$. Thus their navigation behaviors tend to be reactive and previous wrong decisions are hard to be corrected.

To address these limitations, we extend existing \textit{internal memory} based VLN agents with an \textit{external, structured memory}, SSM. Concretely, SSM builds and stores scene layouts and allows easy access to past percepts. SSM also redefines the action space as the set of all the navigable places on the map, allowing making global planning. 
Next, we will first describe the way of constructing SSM (\S\ref{sec:3.1}). Then we introduce iterative reasoning (\S\ref{sec:3.2}) and frontier-exploration based two-step decision making (\S\ref{sec:3.3}), which are two essential components of our SSM based navigator.

\subsection{Structured Scene Memory (SSM) Construction}\label{sec:3.1}
During navigation, our agent online constructs and maintains SSM as an  episodic, topological map of its observed environment. SSM is derived from its past percepts and organized in the form of a graph. Specifically, at time step $t$, SSM is denoted as a directed graph $\mathcal{G}_t\!=\!(\mathcal{V}_t,\mathcal{E}_t)$, where nodes  $v\!\in\!\mathcal{V}_t$ correspond to previously visited places and edges $e_{u,v}\!=\!(u,v)\!\in\!\mathcal{E}_t$ to connections (see Fig.~\ref{fig:overview}(a-b)).  

\noindent\textbf{Visual and Geometric Information Disentangled Layout Representation.}  Each node $v\!\in\!\mathcal{V}^t$ is associated with an embedding $\bm{v}$ to capture visual information at the position it stands for:
\vspace{-1pt}
\begin{equation}\small
\begin{aligned}
\bm{v} \!=\!\{\bm{f}_{v,k}\}_{k=1}^{K_{v}}.
    \end{aligned}
\end{equation}
Here $\bm{v}$ encodes the visual features of the panorama $\{{o}_{v,k}\}_{k=1}^{K_v}$ at position $v$. For
ease of notation, the symbols $\bm{f}_{v,k}$ and $K_{v}$ are slightly reused to represent the $k$-th single-view visual feature and total number of navigable viewpoints at location $v$, respectively.

For each edge $e_{u,v}\!\in\!\mathcal{E}_t$, its embedding $\bm{e}_{u,v}$ encodes geometric cues from position $u$ to $v$:
\vspace{-2pt}
\begin{equation}\small
\begin{aligned}
\bm{e}_{u,v}\!=\!\bm{r}_{u,k_v}\!=\!(\cos\theta_{u,k_v},\sin\theta_{u,k_v}, \cos\phi_{u,k_v}, \sin\phi_{u,k_v}).
    \end{aligned}
    \vspace{-2pt}
\end{equation}
Here $\bm{r}_{u,k_v}$ is slightly reused to represent the orientation feature of $k_v$-th viewpoint at position $u$, from which node $v$ can be navigated. Note that $\bm{e}_{u,v}\neq\bm{e}_{v,u}$.

SSM constructs a topological scene representation from precisely memorized past observations. In addition, visual and geometric cues are disentangled in the node and edge embeddings, respectively. This allows better grounding instructions on the visual world (will be detailed in \S\ref{sec:3.2}).

\noindent\textbf{Global Action Space Modeling.}   Another distinct advantage of SSM is that it provides a global action space. In SSM, each node $v$ is associated with the panoramic view at the corresponding location. Thus $v$ can be divided into $K_v$ sub-nodes $\{{o}_{v,k}\}_{k=1}^{K_v}$ (see Fig.\!~\ref{fig:overview}(c)), and each sub-node represents a navigable viewpoint ${o}_{v,k}$ (\ie, reachable, observable but unvisited).
All the sub-nodes in $\mathcal{G}_{t\!}$ form the action space $\mathcal{A}_{t\!}\!=\!\{{o}_{v,k}\}_{v=1,k=1\!}^{|\mathcal{V}_t|, K_v}$ at step $t$. That means any navigable place on the map can be acted on (\textit{cf.} \S\ref{sec:3.3}).
In contrast, previous VLN agents can only explore currently perceived navigable viewpoints $\{{o}_{t,k}\}_{t}^{K_t\!}$ (Eq.~\ref{eq:decision-making}), \ie, the sub-nodes of current location. In addition, with SSM, the agent can easily change its current navigation direction by directly `jumping' to another sub-node, which may be even observed several steps ago. Thus the agent gets rid of localized backtracking~\cite{ke2019tactical,ma2019regretful} or explore-ahead~\cite{wang2020active} strategies.



\noindent\textbf{Online Updating.} SSM is online built and dynamically updated. At the beginning of an navigation episode, the node set is initialized as the start location. Later, SSM will be extended gradually. Assume that, at step $t$, an action, selected from $\mathcal{A}_t$, is taken to navigate a new place $u$ (\ie, a navigable sub-node). Then a new node will be added to represent $u$. Note that, in $\mathcal{G}_{t}$, there may exist one more sub-nodes correspond to $u$, as a same place can be observed from different viewpoints, and the agent does not know which sub-nodes correspond to $u$ until $u$ is navigated. After visiting $u$, the sub-nodes, in $\mathcal{G}_{t}$, that correspond to $u$ can be found and removed; only the new added node will be left and connected to other navigable nodes. Then, SSM for the next navigation step is constructed, \ie, $\mathcal{G}_{t+1}$.

\subsection{Fine-Grained Instruction Grounding based Iterative Reasoning}\label{sec:3.2}
Given$_{\!}$ current$_{\!}$ navigation$_{\!}$ state,  a$_{\!}$ collect-read$_{\!}$ controller$_{\!}$ is learnt to read useful content from SSM for supporting next-step decision making. It automatically mines perception and object related textual information from instructions, and grounds them onto different elements of SSM. Then it makes use of iterative algorithms for long-range planning.

\noindent\textbf{Perception-and-Action Aware Textual Grounding.} The ability to understand natural-language instructions is critical to building intelligent VLN agents. The instructions interpret navigation routes by linguists, and thus can be grounded in perception and action, instead of arbitrary semantic tokens~\cite{chen2011learning}. For example, instructions describe actions (\eg, \textit{turn left}, \textit{walk forward}) closely relate to orientations, while instructions refer to landmarks in the environment (\eg, \textit{bedroom}, \textit{table}) correspond more to visual perceptions. Whereas visual and orientation information are disentangled in our space representation, we separately explore perception and action related phases and ground them into different SSM elements. Inspired by~\cite{qi2020object}, two different attention models are first learnt to assemble perception and action related textual features from $\bm{X}$, respectively:
\vspace{-2.5pt}
\begin{equation}\small
\begin{aligned}
\!\!\!\!&\bm{x}^{{p}\!}_t\!=\!\text{att}^{{p}\!}(\bm{X}_{\!}, \bm{h}_{t\!})\!=\!\!\sum\nolimits_{l}\!\alpha_{l}^{{p}}\bm{x}_l, ~\!\text{where}~\alpha_{l}^{{p}}\!=\!\text{softmax}_{l}({\bm{x}^{\!\!\top\!}_{l\!}}\bm{W}_{\!x}^{{p}}\bm{h}_{t\!});\!\!\\
\!\!\!\!&\bm{x}^{{a}\!}_t\!=\!\text{att}^{{a}\!}(\bm{X}_{\!}, \bm{h}_{t\!})\!=\!\!\sum\nolimits_{l}\!\alpha_{l}^{{a}}\bm{x}_l, ~\!\text{where}~\alpha_{l}^{{a}}\!=\!\text{softmax}_{l}({\bm{x}^{\!\!\top\!}_{l\!}}\bm{W}_{\!x}^{{a}}\bm{h}_{t\!}).\!\!
\end{aligned}
\label{equ:paattention}
\vspace{-2.5pt}
\end{equation}
Conditioned on current navigation state $\bm{h}_t$, the parsed perception and action related descriptions are of interest to the agent performing next-step navigation.

Next we generate perception and action aware states related to current navigation, respectively:
\vspace{-3.5pt}
\begin{equation}\small
\begin{aligned}
\bm{h}_{t}^{{p}}\!=\!\bm{W}^{{p}}_h[\bm{h}_t, \bm{x}^{{p}}_t];~~~~\bm{h}_{t}^{{a}}\!=\!\bm{W}^{{a}}_h[\bm{h}_t, \bm{x}^{{a}}_t].
\end{aligned}
\label{equ:pastate}
\vspace{-2.5pt}
\end{equation}

Then, for each node $v\!\in\!\mathcal{V}^t$, the collect-read controller assembles the visual
information from its embedding $\bm{v}$, \ie, visual features $\{\bm{f}_{v,k}\}_{k=1}^{K_{v}}$ of the panoramic observations, conditioned on the perception-aware state $\bm{h}_{t}^{{p}}$:
\vspace{-3.5pt}
\begin{equation}\small
\begin{aligned}
\hat{\bm{f}}_v\!=\!\text{att}^{{f}}(\bm{v}, \bm{h}_{t}^{{p}})\!=\!\text{att}^{{f}}(\{\bm{f}_{v,k}\}_{k=1}^{K_{v}}, \bm{h}_{t}^{{p}}).
\end{aligned}
\vspace{-2.5pt}
\label{equ:vass}
\end{equation}


Similarly, for $v\!\in\!\mathcal{V}^t$, the collect-read controller assembles the orientation information from its outgoing edges, \ie, $\{\bm{e}_{v,u}\}_{u\in\mathcal{N}_v}$, conditioned on the action-aware state $\bm{h}_{t}^{\text{a}}$:
\vspace{-6.5pt}
\begin{equation}\small
\begin{aligned}
\hat{\bm{r}}_v\!=\!\text{att}^{{r}}(\{\bm{e}_{v,u}\}_{u\in\mathcal{N}_v}, \bm{h}_{t}^{{a}})\!=\!\text{att}^{{r}}(\{\bm{r}_{v,u}\}_{u\in\mathcal{N}_v}, \bm{h}_{t}^{{a}}),
\end{aligned}
\vspace{-0.5pt}
\label{equ:eass}
\end{equation}
where $\mathcal{N}_{v\!}$ denotes neighbors (\ie, connected locations) of $v$.

\noindent\textbf{Iterative Algorithm based Long-Term Reasoning.} Then we follow classic iterative algorithms to perform long-term reasoning. It is achieved by a parametric \textit{message passing} procedure, which iteratively updates the visual and orientation features by exchanging messages between nodes in the form of trainable
functions. Specifically, at
iteration $s$, $\hat{\bm{f}}_v$ and $\hat{\bm{r}}_v$ of each node $v$ are respectively updated according to the received messages, \ie, $\bm{m}^{f}_v$ and $\bm{m}^{r}_v$, which are information summarized from the neighbors $\mathcal{N}_v$:
\vspace{-3.5pt}
\begin{equation}\small
\begin{aligned}
\bm{m}^{f}_v &=\sum\nolimits_{u\in \mathcal{N}_v}\!\!\hat{\bm{f}}^s_u,~~~~~  \hat{\bm{f}}^{s+1}_v = U^f (\bm{m}^{f}_v, \hat{\bm{f}}^{s}_v);\\
\bm{m}^{r}_v &=\sum\nolimits_{u\in \mathcal{N}_v}\!\!\hat{\bm{r}}^s_u,~~~~~~  \hat{\bm{r}}^{s+1}_v= U^r (\bm{m}^{r}_v, \hat{\bm{r}}^{s}_v),
\end{aligned}
\label{intro1}
\vspace{-2.5pt}	
\end{equation}
where \textit{update functions} $U^*(\cdot)$ are achieved by GRU. After $S$ iterations of aggregation, improved visual and orientation features, \ie, $\hat{\bm{f}}^S_v$  and $\hat{\bm{r}}^S_v$, are generated, through capturing the context within the $S$-hop neighborhood of node $v$.

\vspace{-2pt}
\subsection{Frontier-Exploration based Decision Making}\label{sec:3.3}
\vspace{-2pt}
\noindent\textbf{Global Action Space based One-Step Decision Making.}  SSM provides a global action space $\mathcal{A}_{t\!}\!=\!\{{o}_{v,k}\}_{v=1,k=1\!}^{|\mathcal{V}_t|,K_v}$, comprised of all the navigable sub-nodes in $\mathcal{G}_t$. Each sub-node ${o}_{v,k}$ is associated with the visual ${\bm{f}}_{v,k}$ and orientation ${\bm{r}}_{v,k}$  features of its corresponding viewpoint. We first formulate its confidence of being the next navigation action according to visual and orientation cues, respectively:
\vspace{-4pt}
\begin{equation}\small
\begin{aligned}
q^f_{v,k}\!=\![{\bm{f}}_{v,k}, \hat{\bm{f}}^S_v]^\top\bm{W}^q_f\bm{h}^p_t,~~~~q^r_{v,k}\!=\![{\bm{r}}_{v,k}, \hat{\bm{r}}^S_v]^\top\bm{W}^r_q\bm{h}^a_t.
\end{aligned}
\vspace{-6pt}
\end{equation}
The perception-aware confidence $q^f_{v,k}$ (action-aware confidence $q^r_{v,k}$) is computed by considering the visual (orientation) cues of itself and its parent node. Then the final confidence is computed as a weighted sum of $q^f_{v,k}$ and $q^r_{v,k}$\!~\cite{qi2020object}:
\vspace{-7pt}
\begin{equation}\small
\begin{aligned}
p({o}_{v,k}) = \text{softmax}_{v,k}([q^f_{v,k}, q^r_{v,k}]^\top\bm{w}^p_q),
\end{aligned}
\vspace{-0pt}
\end{equation}
where $\bm{w}^p_q$ is the combination weight: $\bm{w}^p_q=\bm{W}^w[\bm{h}^{{p}}_t, \bm{h}^{{a}}_t]$.

Finally, a feasible navigation action $a_t$ is selected from $\mathcal{A}_{t}$, according to:
\vspace{-8pt}
\begin{equation}\small
\begin{aligned}
~~~~~~~~a_t = \argmax_{v,k} ~p({o}_{v,k}).
\end{aligned}
\vspace{-6pt}
\end{equation}
However, as $\mathcal{A}_{t\!}$ will become large after several navigation steps, this makes policy learning hard and converge slow in practice. To address this issue, we
propose a frontier-exploration based decision making strategy.

\noindent\textbf{Navigation Utility Driven Frontier Selection.} We consider frontiers to be points at the
boundary of known regions of a map~\cite{yamauchi1997frontier}. In $\mathcal{G}_{t}$, the frontiers $\mathcal{W}_{t}\!\subset\!\mathcal{V}_{t}$ are the nodes with more than one navigable sub-nodes, \ie,  $\mathcal{W}_{t}\!=\!\{w|w\!\in\!\mathcal{V}_{t}, K_w\!\geq\!1\}$. In Fig.~\ref{fig:overview}(c), the frontier  nodes are highlighted in green. Frontiers can be viewed important points to facilitate
exploration in a partially observed environment, as no matter which sub-node is selected to be navigated, the navigation must starts at a certain frontier node. Therefore, we define such frontiers as sub-goals for navigation decision making. Instead of directly selecting an action from $\mathcal{A}_{t}$, we first find a frontier node (Fig.~\ref{fig:overview}(d)) and then choose one of its sub-nodes (Fig.~\ref{fig:overview}(e)) to navigate.

At step $t$, a frontier node $w\!\in\!\mathcal{W}_{t}$ is selected according to $\argmax_{w}\!~p(w)$, where $p(w)$ is computed as:
\vspace{-3pt}
\begin{equation}\small
\begin{aligned}
c^f_{w}\!=\!{\hat{\bm{f}}^{S\top}_w}\bm{W}^c_f\bm{h}^p_t,~~~~c^r_{w}\!=\!{\hat{\bm{r}}^{S\top}_w}\bm{W}^c_r\bm{h}^a_t,\\
p(w) = \text{softmax}_{w}([c^f_{w}, c^r_{w}]^\top\bm{w}^p_c).
\end{aligned}
\vspace{-2pt}
\label{equ:fap}
\end{equation}
The agent can easily make error-correction: performing even long-term backtrack by only one-step frontier selection. Our frontier selection policy is learnt in an \textit{navigation utility driven} manner. During RL based online policy learning, the agent is rewarded at each frontier selection step for getting closer to the navigation destination (\textit{cf.} \S\ref{sec:3.4}).  This is different to previous robot control methods selecting frontiers based on certain heuristics~\cite{yamauchi1997frontier,makarenko2002experiment,amigoni2008experimental}, such as
maximizing information gain (the amount of new information that can be acquired) or minimizing moving cost.

\noindent\textbf{Frontier Sub-Node Navigation.} If the intended frontier $w$ is not the current position of the agent, the agent will move to $w$ along the shortest path over $\mathcal{G}_t$. The navigation states, \ie, $\bm{h}_{t}$, $\bm{h}^p_{t}$, and $\bm{h}^a_{t}$, will be also updated by the observations on-the-move. Otherwise, the agent will stay at its current position and maintain current navigation states unchanged. Then, the moving action $a_t$ is selected from the navigable sub-nodes $\{{o}_{w,k}\}_{k=1\!}^{K_w}$ of the selected frontier $w$, according to $\argmax_{k}~p({o}_{w,k})$. Here $p({o}_{w,k})$ is computed by:
\vspace{-3pt}
\begin{equation}\small
\begin{aligned}
q^f_{w,k}\!=\!{\bm{f}}_{w,k}\bm{W}^q_f\bm{h}^p_t,~~~~q^r_{w,k}\!=\!{\bm{r}}_{w,k}\bm{W}^q_r\bm{h}^a_t,\\
p({o}_{w,k}) = \text{softmax}_{k}([q^f_{w,k}, q^r_{w,k}]^\top\bm{w}^p_q).
\end{aligned}
\vspace{-2pt}
\label{equ:nap}
\end{equation}
By performing $a_t$, the$_{\!}$ agent$_{\!}$ will$_{\!}$ reach$_{\!}$ a new position~$u$ (\ie, $\argmax_{w,k}p({o}_{w,k})$). Then, $u$ will be added into SSM, forming $\mathcal{G}_{t+1\!}$ (\textit{cf.} \S\ref{sec:3.1}). The agent will also update its navigation state $\bm{h}_{t}$ with its current panoramic observation $\bm{O}_{t+1}$ (\ie, $\bm{O}_{u}$) through Eq.~\ref{eq:hidden-state}. With the new SSM (\ie, $\mathcal{G}_{t+1}$) and navigation state (\ie, $\bm{h}_{t+1}$), the agent will make next-round navigation. Above process will be executed until the agent thinks it has reached the destination. See \S\ref{sec:abs} for the empirical study of the effectiveness of our frontier-exploration based decision making strategy.



\begin{table*}[t]
\centering
        \resizebox{1\textwidth}{!}{
		\setlength\tabcolsep{8pt}
		\renewcommand\arraystretch{1.0}
\begin{tabular}{c||ccccc|ccccc|ccccc}
\hline \thickhline
\rowcolor{mygray}
~ &  \multicolumn{15}{c}{R2R Dataset} \\
\cline{2-16}
\rowcolor{mygray}
~ &  \multicolumn{5}{c|}{\texttt{validation} \texttt{seen}} & \multicolumn{5}{c|}{\texttt{validation} \texttt{unseen}} & \multicolumn{5}{c}{\texttt{test} \texttt{unseen}} \\
\cline{2-16}
\rowcolor{mygray}
\multirow{-3}{*}{Models} &\textbf{SR}$\uparrow$ &NE$\downarrow$ &TL$\downarrow$ &OR$\uparrow$  &SPL$\uparrow$ &\textbf{SR}$\uparrow$ &NE$\downarrow$ &TL$\downarrow$ &OR$\uparrow$  &SPL$\uparrow$ &\textbf{SR}$\uparrow$ &NE$\downarrow$ &TL$\downarrow$ &OR$\uparrow$ &SPL$\uparrow$\\
\hline
\hline
Student-Forcing~\cite{anderson2018vision} & 0.39 & 6.01 & 11.3 & 0.53  & - & 0.22 & 7.81 & 8.39 & 0.28  & - & 0.20 & 7.85 & \textbf{8.13} & 0.27  & 0.18\\
RPA~\cite{wang2018look} & 0.43 & 5.56 & \textbf{8.46} & 0.53  & - & 0.25 & 7.65 & 7.22  & 0.32  & - & 0.25 & 7.53 & 9.15 & 0.33 & 0.23 \\
E-Dropout~\cite{tan2019learning} & 0.55 & 4.71 & 10.1 & - & 0.53 & 0.47 & 5.49 & 9.37 & - & \textbf{0.43} & - & - & - & -  & - \\
Regretful~\cite{ma2019regretful} & 0.65 & 3.69 & - & 0.72 & \textbf{0.59} & 0.48 & 5.36 & - & 0.61 & 0.37 & - & - & - & - & -\\
EGP~\cite{deng2020evolving} &- & - & - & - & -   &0.52 &5.34  &-  &0.65   &0.41  &  &  &  & \\
Active Perception~\cite{wang2020active} & \textbf{0.66} & \textbf{3.35} & 19.8 & \textbf{0.79} & 0.51 & 0.55 & \textbf{4.40} & 19.9 & \textbf{0.70} & 0.40 & 0.56 & 4.77 & 21.0 & \textbf{0.73} & 0.37\\
\textbf{Ours}& 0.65 & 3.77 & 13.5 & 0.74 & 0.57 & \textbf{0.56} & 4.88 & 18.9 & 0.69 & 0.42 & \textbf{0.57} & \textbf{4.66} & 18.5 & 0.68 & \textbf{0.44} \\
\hline
Speaker-Follower~\cite{fried2018speaker}* & 0.66 & 3.36 & - & 0.74 & - & 0.36 & 6.62 & - & 0.45  & - & 0.35 & 6.62 & 14.8 & 0.44 & 0.28\\
RCM~\cite{wang2019reinforced}*& 0.67 & 3.53 & \textbf{10.7} & 0.75 & - & 0.43 & 6.09 & 11.5 & 0.50 & - & 0.43 & 6.12 & 12.0 & 0.50 & 0.38 \\
Self-Monitoring~\cite{ma2019self}*& 0.67 & 3.22 & - & 0.78 & 0.58 & 0.45 & 5.52 & - & 0.56 & 0.32 & 0.43 & 5.99 & 18.0 & 0.55 & 0.32 \\
Regretful~\cite{ma2019regretful}* & 0.69 & 3.23 & - & 0.77  & 0.63 & 0.50 & 5.32 & - & 0.59  & 0.41 & 0.48 & 5.69 & 13.7 & 0.56  & 0.40 \\
E-Dropout~\cite{tan2019learning}* &0.62 & 3.99 & 11.0 & - & 0.59 & 0.52 & 5.22 & \textbf{10.7} & - &0.48 & 0.51 & 5.23 & \textbf{11.7} & 0.59  & 0.47 \\
OAAM*~\cite{qi2020object} &0.65 &-  &10.2  &0.73  &0.62  &0.54 &- &9.95  &0.61  &0.50   &0.53  &-  &10.4  &0.61  &0.50 \\
EGP*~\cite{deng2020evolving}& - & - & - & - & -  &0.56 &4.83 &-  &0.64  &0.44   &0.53  &5.34  &-  &0.61  &0.42 \\
Tactical Rewind~\cite{ke2019tactical}*& - & - & - & - & -&0.56 & 4.97  & 21.2 & - &0.43 & 0.54 & 5.14 & 22.1 & 0.64 & 0.41 \\
AuxRN~\cite{zhu2019vision}*&0.70 &3.33 & - & 0.78 & \textbf{0.67} &0.55 &5.28 & - & 0.62 & \textbf{0.50}& 0.55 &5.15 & - &0.62 & \textbf{0.51} \\
Active Perception~\cite{wang2020active}*& 0.70 & 3.20 & 19.7 &\textbf{0.80} & 0.52 & 0.58 & 4.36 & 20.6 & 0.70 & 0.40  & 0.60 & \textbf{4.33} & 21.6 & \textbf{0.71}  & 0.41 \\
\textbf{Ours}*&\textbf{0.71}&\textbf{3.10}&14.7&\textbf{0.80}&0.62&\textbf{0.62}&\textbf{4.32}&20.7&\textbf{0.73}&0.45& \textbf{0.61} & 4.57 & 20.4 & 0.70 & 0.46 \\
\hline
\end{tabular}
}
	\vspace*{-4pt}
\captionsetup{font=small}
	\caption{\small{Quantitative comparison results (\S\ref{sec:R2R}) on R2R dataset~\cite{anderson2018vision}.  For compliance with the evaluation server, we report SR as fractions.  `$^*$': back translation based data augmentation. `$-$': unavailable statistics.}}
    \label{table:R2R}
\vspace*{-7pt}
\end{table*}

\subsection{Implementation Details}\label{sec:3.4}
\noindent\textbf{Network Architecture:}  We build upon a basic navigation architecture in~\cite{tan2019learning}. As in~\cite{anderson2018vision,wang2019reinforced,wang2018look}, each panoramic view is divided into 36 sub-views, \ie, 12\!~headings\!~$\times$\!~3\!~elevations with 30 degree intervals.  For each viewpoint $o_{t,k}$, the visual embedding $\bm{f}_{t,k}$ is a 2048-$d$ ResNet-152~\cite{he2016deep} feature and orientation embedding $\bm{r}_{t,k}$ is a 128-$d$ vector (\ie, 4-$d$ orientation feature $(\cos\theta_{t,k}, \sin\theta_{t,k}, \cos\phi_{t,k}, \sin\phi_{t,k})$ are tiled 32 times). Instruction embeddings $\bm{{X}}_{\!}$ are obtained from an LSTM with a 512 hidden size.

\begin{figure*}[t]
	\begin{center}
		\includegraphics[width=0.9\linewidth]{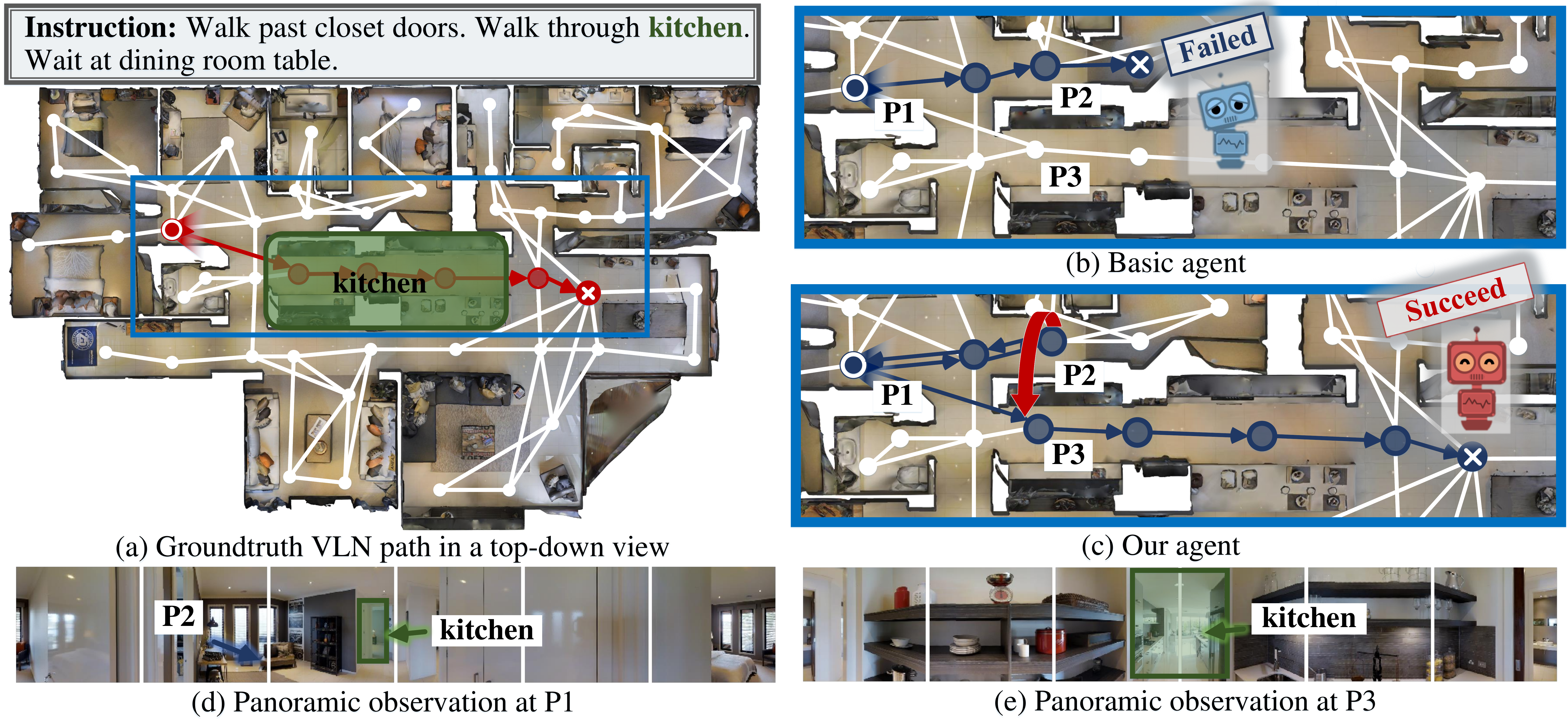}
	\end{center}
	\vspace{-16pt}
	\captionsetup{font=small}
	\caption{\small{A representative visual result on R2R dataset~\cite{anderson2018vision} (\S\ref{sec:R2R}). Due to partial observability of agent's perception system, it is hard to find kitchen at position P1 (see (d)). Thus the basic agent, in (b), goes the wrong way and ends in failure. However, in (c), our agent easily turns back and explores another direction (\ie, P3). At P3, the kitchen is easy to find (see (e)) and the instruction is successfully completed.} }
	\label{fig:vr2r}
	\vspace{-12pt}
\end{figure*}

\noindent\textbf{Network Training:} Following~\cite{zhu2017target,wang2018look,tan2019learning,wang2019reinforced}, our model is trained using both imitation learning  (IL) and reinforcement learning (RL). For IL, we adopt both teacher forcing and student forcing to train our agent. In teacher forcing, the agent learns the offline policy through behavior cloning of the expert navigation trajectories. However, the agent is biased to the teacher action and limits exploration to a local action space that is the in
ground-truth shortest-path trajectory. In student forcing, the agent samples its next-step action from its predicted probability distribution. We choose the node on SSM that is closest to the target location as the teacher action.  In RL based online policy learning, the rewards for frontier and action selection are designed in a similar spirit. At each frontier or navigation action selection step, the agent receives an immediate reward, \ie, the change in the distance to the goal destination. At the end of the episode, the agent receives a reward only if it terminated successfully~\cite{wang2019reinforced,tan2019learning}. In addition, following~\cite{fried2018speaker,tan2019learning,qi2020object,deng2020evolving,ke2019tactical}, back translation~\cite{fried2018speaker} based data augmentation technique is used for  R2R dataset~\cite{anderson2018vision}.

\noindent\textbf{Reproducibility:} Our model is implemented in PyTorch and trained on 8 NVIDIA Tesla V100 GPUs. To reveal full details of our method, our implementations are released.

\vspace{-4pt}
\section{Experiment}
\vspace{-4pt}
\subsection{Performance on R2R Dataset}\label{sec:R2R}
\vspace{-3pt}
\noindent\textbf{Dataset.} We first conduct experiments on R2R dataset~\cite{anderson2018vision} which is split into four sets: \texttt{training} ($61$ environments, $14,039$ instructions), \texttt{validation} \texttt{seen} ($61$ environments, $1,021$
instructions), \texttt{validation} \texttt{unseen} ($11$ environments, $2,349$ instructions), and \texttt{test} \texttt{unseen} ($18$ environments, $4,173$ instructions). There are no overlapping environments between the unseen and training sets.


\begin{table*}[t]
    \centering
            \resizebox{1\textwidth}{!}{
            \setlength\tabcolsep{7pt}
            \renewcommand\arraystretch{1.0}
    \begin{tabular}{c||cccccc|cccccc}
    \hline \thickhline
    \rowcolor{mygray}
    ~ &  \multicolumn{12}{c}{R4R Dataset} \\
    \cline{2-13}
    \rowcolor{mygray}
    ~ & \multicolumn{6}{c|}{\texttt{validation} \texttt{seen}}& \multicolumn{6}{c}{\texttt{validation} \texttt{unseen}} \\
    \cline{2-13}
    \rowcolor{mygray}
    \multirow{-3}{*}{Models} &NE$\downarrow$ &TL$\downarrow$ &SR$\uparrow$ & CLS$\uparrow$ & nDTW$\uparrow$ & SDTW$\uparrow$ &NE$\downarrow$ &TL$\downarrow$ &SR$\uparrow$ & CLS$\uparrow$ & nDTW$\uparrow$ & SDTW$\uparrow$\\
    \hline
    \hline
    Speaker-Follower~\cite{fried2018speaker} & 5.35 & 15.4  & 0.52 & 0.46 & - & - & 8.47 & 19.9  & 0.24 & 0.30 & - & - \\
    RCM+goal oriented~\cite{wang2019reinforced}& 5.11 & 24.5  & 0.56 & 0.40 & - & - & 8.45 & 32.5  & 0.29 & 0.20 & 0.27 & 0.11 \\
    RCM+fidelity oriented~\cite{wang2019reinforced} & 5.37 & 18.8  & 0.53 & 0.55 & - & - & 8.08 & 28.5  & 0.26 & 0.35 & 0.30 & 0.13\\
    PTA low-level~\cite{landi2020perceive} & 5.11 & 11.9  & 0.57 & 0.52 & 0.42 & 0.29 & 8.19 & \textbf{10.2}  & 0.27 & 0.35 & 0.20 & 0.08 \\
    PTA high-level~\cite{landi2020perceive} & \textbf{4.54} & 16.5  & 0.58 & 0.60 & \textbf{0.58} & 0.41 & 8.25 & 17.7  & 0.24 & 0.37 & 0.32 & 0.10\\
    EGP~\cite{deng2020evolving} & - & -  & - & - & - & - & \textbf{8.00} & 18.3  & 0.30 & 0.44 & 0.37 & 0.18\\
    E-Drop~\cite{tan2019learning} & - & 19.9  & 0.52 & 0.53 & - & 0.27 & - & 27.0  & 0.29 & 0.34 & - & 0.09 \\
    OAAM~\cite{qi2020object} & - & \textbf{11.8}  & 0.56 & 0.54 & - & 0.32 & - & 13.8  & 0.31 & 0.40 & -  & 0.11\\
    \textbf{Ours} &4.60 &19.4 &\textbf{0.63} &\textbf{0.65} & 0.56 &\textbf{0.44} &8.27 &22.1 &\textbf{0.32} &\textbf{0.53} &\textbf{0.39} &\textbf{0.19}\\
    \hline
    \end{tabular}
    }
        \vspace*{-4pt}
    \captionsetup{font=small}
	\caption{\small{Quantitative comparison results (\S\ref{sec:R4R}) on  R4R  dataset~\cite{jain2019stay}. `$-$': unavailable statistics.}}    \label{table:R4R}
    \vspace*{-7pt}
    \end{table*}

\begin{figure*}[t]
	\begin{center}
		\includegraphics[width=0.99\linewidth]{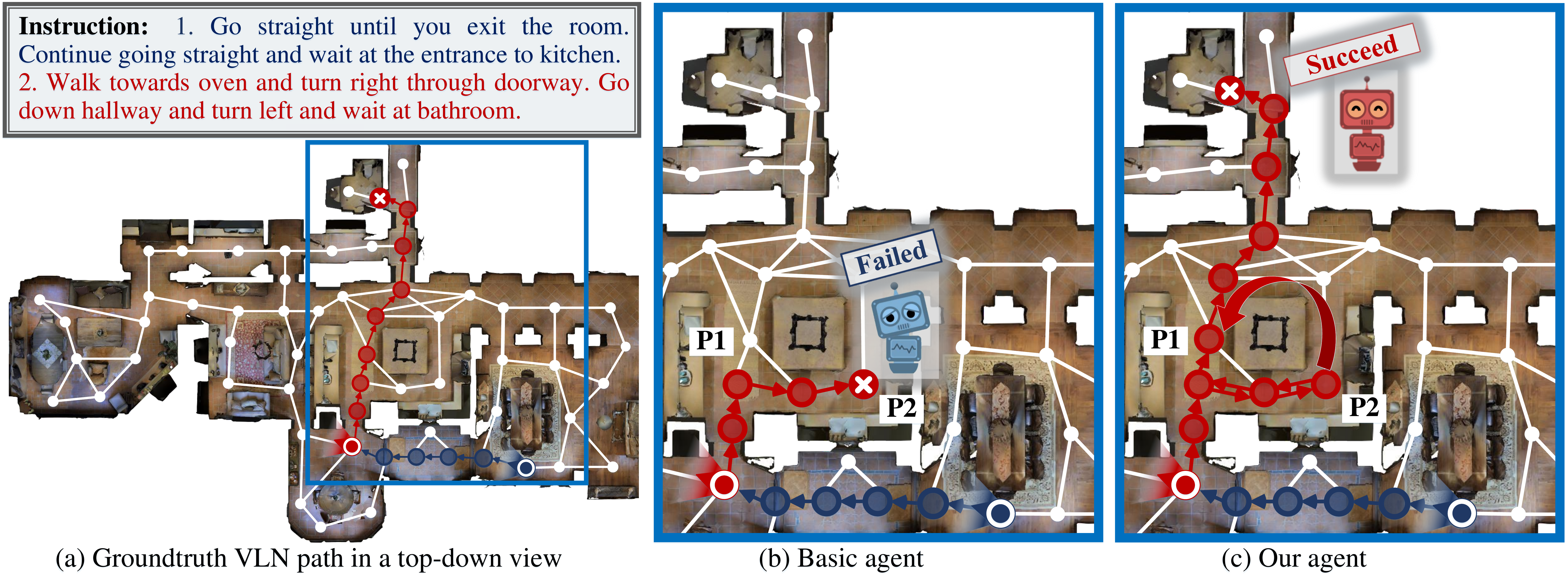}
	\end{center}
	\vspace{-15pt}
	\captionsetup{font=small}
	\caption{\small{A representative visual result on R4R dataset~\cite{jain2019stay} (\S\ref{sec:R4R}). The instruction ``\textcolor{bblue}{\textit{Go straight} $\cdots$}'' is easy to follow and corresponding navigation trajectories are highlighted in blue. However, in (b), the basic agent is confused by the ambiguous instruction ``\textcolor{rred}{$\cdots$\!~\textit{turn right through doorway}\!~$\cdots$}'', causing failed navigation. Through SSM, our agent can easily access to past observations and make more robust navigation. As shown in (c), our agent successfully returns to the correct direction and reaches the target location finally.} }
	\label{fig:vr4r}
	\vspace{-12pt}
\end{figure*}

\noindent\textbf{Evaluation Metric.} Following~\cite{anderson2018vision,fried2018speaker}, five metrics are used:
(1) \textit{Success Rate} (SR), the primary metric, considers the percentage of final positions less than 3 m away from the goal location.
(2) \textit{Navigation Error} (NE) refers to the shortest distance between agent's final position and the goal location.
(3) \textit{Trajectory Length} (TL)  measures the total length of agent trajectories.
(4) \textit{Oracle success Rate} (OR) is the success rate if the agent can stop at the closest point to the goal along its trajectory.
(5) \textit{Success rate weighted by Path Length} (SPL)~\cite{anderson2018evaluation}  is a trade-off between  SR and TL.

\noindent\textbf{Quantitative Result.} Table~\ref{table:R2R} compares our model against 12 state-of-the-art VLN models on R2R dataset. We find that our model outperforms other competitors across most metrics.  For instance, compared with current top-leading method, Active Perception~\cite{wang2020active}, our model performs better in term of SR, and achieves a huge boost in SPL, across \texttt{validation} \texttt{seen}, \texttt{validation} \texttt{unseen}, and \texttt{test} \texttt{unseen}. Our model also significantly surpasses the second best method, AuxRN~\cite{zhu2019vision}, in terms of SR, NE, TL and OR, without using any auxiliary supervision signals. Despite having similar motivation of exploring global, structured action space, EGP~\cite{deng2020evolving} achieves an SR  of 0.53 on \texttt{test} \texttt{unseen} while our model yields 0.61. We also notice that our agent without data augmentation already outperforms existing methods on SR, NE and OR, only excepting~\cite{wang2020active}.

\noindent\textbf{Visual Result.} As shown in Fig.~\ref{fig:vr2r}(a), \textit{kitchen} is a critical landmark for instruction execution. However, at the start location (\ie, P1), \textit{kitchen} is hard to be observed (Fig.~\ref{fig:vr2r}(d)). Thus the basic agent navigates a wrong direction and finally fails (Fig.~\ref{fig:vr2r}(b)). However, with the help of SSM, our agent can make long-range planning over all the explored environment and quickly shift its navigation direction from P2 to P3 (Fig.~\ref{fig:vr2r}(c)). As shown in Fig.~\ref{fig:vr2r}(e), our agent then finds \textit{kitchen} and accomplishes the instruction successfully.

\vspace{-4pt}
\subsection{Performance on R4R Dataset}\label{sec:R4R}
	\vspace{-2pt}
\noindent\textbf{Dataset.} R4R~\cite{jain2019stay} is  an
extended version of R2R with longer instructions and trajectories. R4R has
$278,766$ instructions and contains three splits: \texttt{training} ($61$ environments, $233,613$ instructions), \texttt{validation} \texttt{seen} ($61$ environments, $1,035$
instructions), and \texttt{validation} \texttt{unseen} ($11$ environments, $45,162$ instructions).

\begin{table*}[t]
    \begin{center}
            \resizebox{1\textwidth}{!}{
            \setlength\tabcolsep{2pt}
            \renewcommand\arraystretch{1.05}
    \begin{tabular}{c|c||ccccc|ccccc}
    \hline \thickhline
    \rowcolor{mygray}
    & & \multicolumn{10}{c}{R2R Dataset} \\
    \cline{3-12}
    \rowcolor{mygray}
    & & \multicolumn{5}{c|}{\texttt{validation} \texttt{seen}} & \multicolumn{5}{c}{\texttt{validation} \texttt{unseen}} \\
    \cline{3-12}
    \rowcolor{mygray}
    \multicolumn{1}{c|}{\multirow{-3}{*}{Model}} &\multirow{-3}{*}{SSM Component} &{SR}$\uparrow$ &NE$\downarrow$ &TL$\downarrow$ &OR$\uparrow$ &SPL$\uparrow$ &{SR}$\uparrow$ &NE$\downarrow$ &TL$\downarrow$ &OR$\uparrow$  &SPL$\uparrow$\\
    \hline
    \hline
    Basic agent~\cite{tan2019learning} &- &0.62 & 3.99 & 11.0 & 0.71 & 0.59 & 0.52 & 5.22 & 10.7 & 0.58 &0.48 \\
    \hline
    \multirow{2}{*}{Full model} &\textbf{Fine-grained instruction grounding} + \textbf{Iterative reasoning}& \multirow{2}{*}{{0.71}}&\multirow{2}{*}{{3.10}}&\multirow{2}{*}{14.7}&\multirow{2}{*}{0.80}&\multirow{2}{*}{0.62} &\multirow{2}{*}{{0.62}}&\multirow{2}{*}{{4.32}}&\multirow{2}{*}{20.7}&\multirow{2}{*}{0.73}&\multirow{2}{*}{0.45} \\
    \specialrule{0em}{-0.5pt}{-1.5pt}
    &+ \textbf{Frontier-exploration based decision making}&&&&&&&&&& \\
    \hline
    &Full model \textbf{\textit{w/o.} fine-grained instruction grounding} & 0.70 & 3.44 & 13.5 & 0.80 & 0.62 & 0.59 & 4.57 & 20.3 & 0.72 & 0.43 \\
    &Full model \textbf{\textit{w/o.} iterative reasoning} & 0.69 & 3.51 & 12.9 & 0.78 & 0.61 & 0.57 & 4.60 & 19.8 & 0.70 & 0.42 \\
    &Full model \textbf{\textit{w.} local action space based decision making} & 0.67 & 3.62 & 11.9 & 0.75 & 0.63 & 0.54 & 5.21 & 12.6 & 0.64 & 0.49\\
    \multirow{-4}{*}{Variants}&Full model \textbf{\textit{w.} global action space based one-step decision making} &0.68 &3.25& 13.2& 0.74& 0.60& 0.56 & 4.76 & 19.8 &0.68 & 0.41\\
    \hline
    \end{tabular}
    }
    \end{center}
        \vspace*{-12pt}
    \captionsetup{font=small}
	\caption{\small{Ablation study (\S\ref{sec:abs}) on \texttt{validation} \texttt{seen} and \texttt{validation} \texttt{unseen} sets of R2R dataset~\cite{anderson2018vision}.}}\label{table:abs}
    \vspace*{-13pt}
\end{table*}

\noindent\textbf{Evaluation Metric.}   Following~\cite{wang2019reinforced,landi2020perceive,deng2020evolving}, six evaluation metrics are used for R4R. Specifically, in addition to (1) SR, (2) NE, and (3) TL, three new metrics are introduced: (4) \textit{Coverage weighted by Length Score} (CLS)~\cite{jain2019stay} that considers both how well the ground truth path is covered by the agent trajectory and the trajectory length, (5) \textit{normalized Dynamic Time Warping} (nDTW) that measures the fidelity (order consistency) of agent trajectories, and (6) \textit{Success rate weighted normalized
Dynamic Time Warping} (SDTW) that stresses the importance of reaching the goal comparing to nDTW. Here CLS is considered as the primary metric.

\noindent\textbf{Quantitative Result.} Table~\ref{table:R4R} presents comparisons with six top-leading VLN models on R4R dataset. We can find that our approach sets new state-of-the-arts for most metrics. Notably, on \texttt{validation} \texttt{seen}  and \texttt{validation} \texttt{unseen}, our model
yields CLSs of 0.65 and 0.53, respectively, while those for the second best method are 0.60 and 0.44. Our model also shows significant performance gains in terms of SR, nDTW, and SDTW. In addition, among all the methods, our agent achieves smallest performance gap between seen and unseen environments, \eg, 0.12 in terms of CLS. This evidences the advantage of SSM that helps our agent learn more generalized navigation policy. Note that~\cite{hong2020language} is excluded from the comparisons on R2R and R4R, as it makes use of extra data and object-level cues.

\noindent\textbf{Visual Result.}  Fig.~\ref{fig:vr4r}(a) gives a challenging case in  R4R. There are two conjuncted instructions, corresponding to twisted routes connecting two shortest-path trajectories in R2R, are highlighted in blue and red, respectively. After successfully accomplishing the first instruction ``\textit{Go straight}\!~$\cdots$", the basic agent is misled by the ambiguous instruction ``$\cdots$\!~\textit{turn right through doorway}\!~$\cdots$" and mistakenly stops at position P2 (Fig.~\ref{fig:vr4r}(b)). However, SSM allows our agent to make more robust decision on the global action space. As shown in Fig.~\ref{fig:vr4r}(c), our agent easily returns back to the correction direction (\ie, P1) and finally succeeds.

\subsection{Diagnostic Experiments}
\label{sec:abs}

For thoroughly assessing the efficacy of essential components of our approach, a set of ablation studies are conducted on the two validations sets of  R2R dataset~\cite{anderson2018vision}.


\noindent\textbf{Whole SSM Architecture Design.} As shown in the first two rows in Table~\ref{table:abs}, our ``Full model'' significantly outperforms the ``Basic Agent''~\cite{tan2019learning} on both \texttt{validation} \texttt{seen} and \texttt{validation} \texttt{unseen} sets, across all metrics. This demonstrates the effectiveness of our whole model design.

\noindent\textbf{Perception-and-Action Aware Instruction Grounding.} In SSM, visual and geometric information can be disentangled (\S\ref{sec:3.1}). This powerful scene
representation allows our agent to better interpret instructions and predict more accurate navigation actions  (\S\ref{sec:3.2}). To verify this point, we only use the original, perception-and-action agnostic navigation state $\bm{h}_{t}$, instead of perception and action aware states (\ie, $\bm{h}_{t}^{{p}}$ and $\bm{h}_{t}^{{a}}$ in Eq.~\ref{equ:pastate}), for visual and geometric information assembling (Eqs.\!~\ref{equ:vass}-\ref{equ:eass}) and action probability estimation (Eqs.~\ref{equ:fap}-\ref{equ:nap}). Considering the second and third rows in Table~\ref{table:abs}, we can conclude that visual and geometric information disentangled scene representation with fine-grained instruction grounding indeed boosts VLN performance.

\noindent\textbf{Iterative Reasoning.} In \S\ref{sec:3.2}, a parametric message passing procedure is conducted over our SSM, which step-by-step gathers information for long-range planning and decision-making supporting. After omitting this procedure, we obtain a baseline ``Full model \textbf{\textit{w/o.} iterative reasoning}''. Comparing this baseline with our ``Full model'', significant performance drop can be observed (see Table~\ref{table:abs}).

\noindent\textbf{Frontier-Exploration based Decision Making.} Our SSM provides a global navigation action space -- the agent is able to navigate any past visited location when necessary (\S\ref{sec:3.3}). To demonstrate this advantage, we develop a new baseline ``Full model \textbf{\textit{w.} local action space based decision making}'', \ie, the agent is only allowed to navigate current visible directions. This strategy is widely adopted in current VLN models. As evidenced in Table~\ref{table:abs}, the performance of the new baseline is significantly worse than ``Full model', revealing the necessity of considering a global action space. To erase the difficulty of policy learning on the fast-growing global action space, we develop a frontier-exploration based decision making strategy. Compared with directly making decision on the whole global action space (\ie, ``Full model \textbf{\textit{w.} global action space based one-step decision making}''), our strategy obtains a consistent margin across all metrics.

\section{Conclusion}

Memory and mapping are crucial components to enabling intelligent navigation in partially observable environments. However, current VLN agents are simply built upon recurrent neural networks. Their latent network state based implicit memory is unsuitable for modeling long-term structural data dependencies, hence limiting their planning ability. In this paper, we develop a structured, explicit memory architecture, SSM, that allows VLN agents to access to its past percepts and explore environment layouts.  With this expressive and persistent space representation, our agent shows advantages in fine-grained instruction grounding, long-term reasoning, and global decision-making. We demonstrate empirically that our SSM based VLN agent
sets state-of-the-arts on challenging R2R and R4R datasets.

%

%

{\small
\bibliographystyle{ieee_fullname}
\bibliography{egbib}
}

\end{document}